\documentclass[letterpaper, 10 pt, conference]{ieeeconf}  

\usepackage{graphicx}
\usepackage{amsmath}
\usepackage{amssymb}
\usepackage{multirow}
\usepackage{makecell}
\usepackage{subfig}
\usepackage{svg}
\usepackage{threeparttable}
\IEEEoverridecommandlockouts                              

\title{\LARGE \bf
A2DO: Adaptive Anti-Degradation Odometry with Deep Multi-Sensor Fusion for Autonomous Navigation
}

\author{Hui Lai$^{\dagger}$ , Qi Chen$^{\dagger}$, Junping Zhang, Jian Pu$^{*}$
\thanks{$^{*}$ Corresponding author}
\thanks{${\dagger}$ These two authors.contribute equally to this work}%
\thanks{All authors are with Fudan University, Shanghai 200433, China}
\thanks{\{21262010024, qichen21\}@m.fudan.edu.cn, }
\thanks{\{jpzhang, jianpu\}@fudan.edu.cn.}
}

\begin{document}

\maketitle
\thispagestyle{empty}
\pagestyle{empty}

\begin{abstract}
Accurate localization is essential for the safe and effective navigation of autonomous vehicles, and Simultaneous Localization and Mapping (SLAM) is a cornerstone technology in this context. However, The performance of the SLAM system can deteriorate under challenging conditions such as low light, adverse weather, or obstructions due to sensor degradation. We present A2DO, a novel end-to-end multi-sensor fusion odometry system that enhances robustness in these scenarios through deep neural networks. A2DO integrates LiDAR and visual data, employing a multi-layer, multi-scale feature encoding module augmented by an attention mechanism to mitigate sensor degradation dynamically. The system is pre-trained extensively on simulated datasets covering a broad range of degradation scenarios and fine-tuned on a curated set of real-world data, ensuring robust adaptation to complex scenarios. Our experiments demonstrate that A2DO maintains superior localization accuracy and robustness across various degradation conditions, showcasing its potential for practical implementation in autonomous vehicle systems.

\end{abstract}

\section{INTRODUCTION}
The advent of autonomous vehicles heralds a new era in intelligent transportation systems, promising enhanced mobility and safety. Central to this promise is the ability to achieve real-time, precise localization, which is crucial for navigation and collision avoidance. Odometry stands out as a pivotal technology that empowers vehicles to determine their position and construct a map of the environment in real-time, without the need for pre-existing maps \cite{jinke2022review}. Despite its potential, traditional odometry systems often struggle to maintain localization accuracy under challenging conditions such as low-light scenarios, inclement weather, or obstructions. These scenarios underscore the pressing need for more robust SLAM solutions that can reliably operate under diverse real-world conditions.

Multi-sensor fusion effectively addresses sensor degradation by combining data from complementary sensors, including cameras, LiDARs, and IMUs. Individual sensors may fail under specific conditions, such as LiDAR in rainy scenarios, cameras in low-light scenarios, and IMUs suffering from drift fusion. Previous geometric-based methods such as 
\cite{chen2024multi}, \cite{chen2024vpl}
perform well in various scenarios. However, the reliance on rule-based approaches\cite{zhang2016degeneracy} for degraded sensor data makes these systems less effective in complex scenarios and requires significant manual calibration and tuning. Deep learning-based methods show great potential in odometry tasks \cite{chen2020survey}, excelling in sparse features and dynamic scenarios. These methods demonstrate increased robustness in degraded conditions, offering flexibility in feature fusion and reducing sensitivity to calibration and synchronization. However, these methods typically require extensive real-world data for training, and their performance in complex degraded scenarios often hinges on the availability of such data. Collecting real-world data in challenging conditions remains difficult\cite{han2023carla}, limiting their practical application.

To address the challenges inherent in multi-sensor fusion odometry, we present a novel, robust, multi-sensor fused odometry system that integrates deep learning techniques. The proposed system employs deep neural networks (DNNs) to develop an end-to-end odometry framework that adaptively mitigates the effects of sensor degradation. By leveraging the advanced feature extraction capabilities of DNNs, the system overcomes the limitations of traditional feature-based methods, especially in degradation and dynamic scenarios. Our system is extensively pre-trained on simulated datasets containing diverse degradation scenarios, facilitating effective transfer to real-world driving scenarios with minimal reliance on large-scale real-world data. This approach ensures high localization accuracy and robustness even in challenging degraded conditions. The primary contributions of this work are as follows:

\begin{itemize}
    \item We propose A2DO, an end-to-end multi-sensor fusion odometry system endowed with adaptive degradation handling capabilities. Through comprehensive evaluations in complex autonomous driving scenarios, we demonstrate that the proposed system consistently maintains high localization accuracy and robustness across various degradation conditions.
    \item Our multi-layer, multi-scale feature encoding module can effectively integrate LiDAR and visual data. By incorporating an attention mechanism within the high-dimensional latent feature space, the system sequentially filters temporal and spatial features, thereby enhancing the efficiency of feature fusion and improving the system's stability in complex scenarios.
    \item Our system undergoes extensive pre-training on simulated datasets featuring a wide range of degradation scenarios, followed by fine-tuning the model on a small set of real-world data. This training regimen enables efficient transfer to diverse driving scenarios, ensuring robust and accurate localization in real-world complex scenarios, thereby validating the practical applicability of the proposed odometry system.
\end{itemize}

\section{RELATED WORKS}

\subsection{Traditional Geometric-Based Methods}
Traditional multi-sensor fusion odometry systems built on geometric principles have established robust theoretical foundations. These approaches can be broadly categorized into filter-based and optimization-based methods.
Filter-based methods proposed in 
\cite{chen2024multi}
\cite{lin2022r}, 
utilize the Extended Kalman Filter (EKF), fuse IMU data with external sensors like cameras or LiDAR to update the vehicle's state and improve localization accuracy. 
Notably, Multi-LIO \cite{chen2024multi} seamlessly integrates multiple LiDARs with an IMU to deliver robust odometry, while R3LIVE \cite{lin2022r} builds upon LiDAR-inertial frameworks \cite{xu2022fast} by incorporating photometric errors from visual data, thereby improving both accuracy and resilience.
However, these methods 
lack specific mechanisms to address sensor degradation under extreme conditions.
Optimization-based methods such as \cite{shan2021lvi}, \cite{zhao2021super} utilize pose graph and factor graph optimization\cite{labbe2014online}, treat states and sensor parameters as nodes, while residuals form the edges. 
Methods such as VPL-SLAM \cite{chen2024vpl} and UL-SLAM \cite{jiang2024ul} leverage visual line features to enhance the robustness of traditional visual SLAM systems\cite{campos2021orb}, while Super Odometry \cite{zhao2021super} adopts a loosely-coupled architecture to maintain flexibility under sensor degradation.
However, these approaches often rely on rule-based handling\cite{zhang2016degeneracy} of degraded sensor data, making them less effective in complex scenarios, and they typically require extensive manual calibration and parameter tuning.

\subsection{End-to-End Deep Learning Methods}
Recent advances in deep learning have led to data-driven, end-to-end multi-sensor fusion methods. Deeplio\cite{iwaszczuk2021deeplio} converts LiDAR point clouds into 2D vertex and normal images, using CNNs and RNNs to fuse LiDAR and IMU data in a deep learning framework for localization.
Wang et al.\cite{wang2022attention} introduced an attention-based visual-inertial odometry system, where IMU-derived motion states query CNN-extracted visual depth and optical flow features. Other methods, such as Selectfusion\cite{chen2019selectfusion} and Yang's efficient fusion strategy\cite{yang2022efficient}, enhance robustness by reweighting high-dimensional features using soft mask attention mechanisms. TransFusionOdom\cite{sun2023transfusionodom} further innovates by transforming both LiDAR and IMU data into 2D images, leveraging ResNet\cite{he2016deep} and Transformer\cite{transformer} architectures to achieve precise 6-DoF pose estimation. These deep learning methods offer greater robustness in degraded scenarios and exhibit higher flexibility in feature fusion while reducing sensitivity to sensor calibration and synchronization issues.
However, the efficacy of existing deep learning approaches frequently hinges on their performance within a singular dataset, particularly in their capacity to handle degradation scenarios, thereby still presenting challenges in terms of generalization.

\section{METHOD}
\begin{figure*}[ht]
    \centering
    \includegraphics[width=\linewidth]{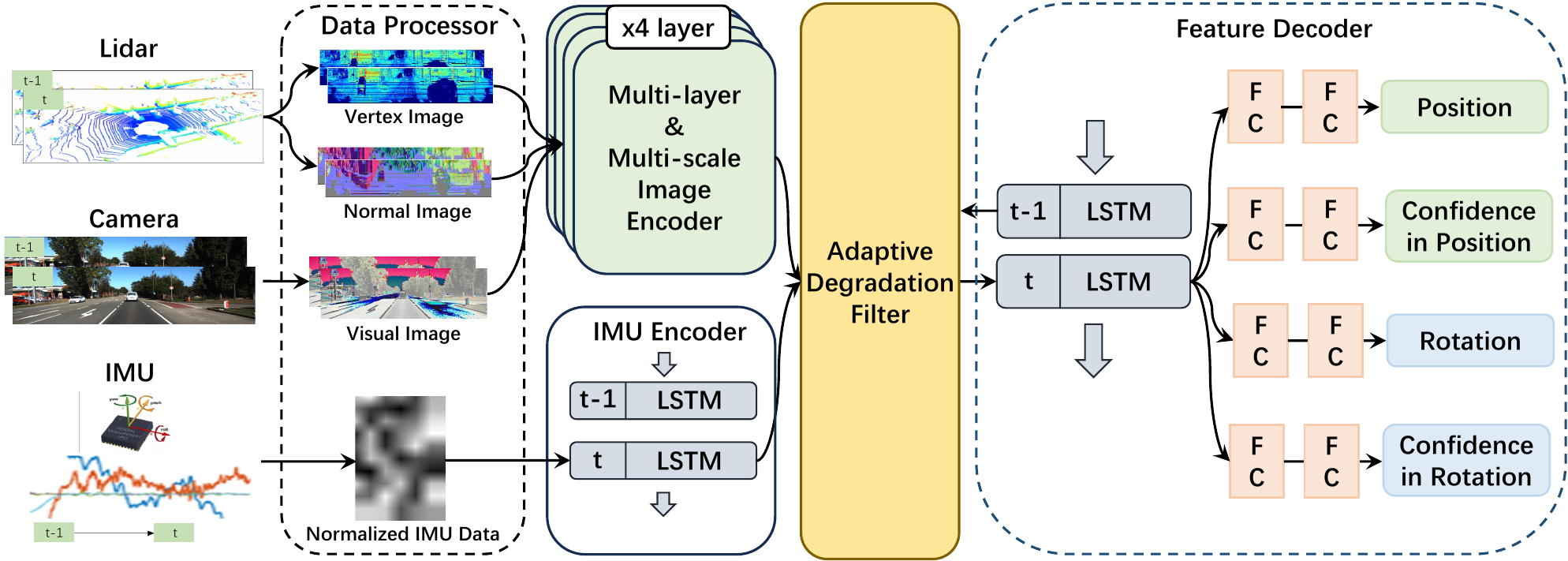}
    \caption{A2DO framework pipeline. Raw sensor (LiDAR, Camera, IMU) data is preprocessed via 2D projection and timestamp alignment. The processed vertex, normal, and visual images are encoded by a multi-layer and multi-scale ResNet-Transformer, while normalized IMU data is handled by a lightweight LSTM. Latent features are refined through an adaptive degradation filter. Finally, an LSTM-based decoder estimates the 6-DOF vehicle pose with corresponding confidence scores.}
    \label{fig:do_arch}
\end{figure*}

As illustrated in Fig.\ref{fig:do_arch}, our system follows an encoder-decoder architecture with adaptive hierarchical filtering applied to latent features for efficient sensor data fusion. The final output includes the 6-DOF (Degrees of Freedom) vehicle pose and corresponding confidence scores. The key components are as follows:
\begin{itemize}
    \item Data Processor: Once the system receives the point cloud frame from LiDAR, we transform these points into vertex and normal images using spherical projection method\cite{milioto2019rangenet++}, while RGB and IMU data are timestamp-aligned, normalized, and stacked for encoder input.
    \item Feature Encoder: We design a multi-layer, multi-scale encoder combining ResNet and Transformer architectures to extract and fuse LiDAR and Camera features efficiently. We apply a lightweight LSTM-based encoder for low-dimensional IMU data to capture temporal dependencies.
     \item Adaptive Degradation Feature Filter: To deal with degraded features, we design a coarse-to-fine filtering strategy on encoded latent features, which includes both a Temporal Feature Filter and a Spatial Feature Filter, ensuring the odometry against various degradation scenarios.
     
    \item Feature Decoder: An LSTM-based decoder fuses filtered features to estimate the 6-DOF vehicle pose and provides historical state information for adaptive temporal filtering.
\end{itemize}

\subsection{Multi-layer and Multi-scale Image Encoder} \label{MM_encoder} 
\begin{figure}[]
    \centering
     \includegraphics[width=0.85\linewidth]{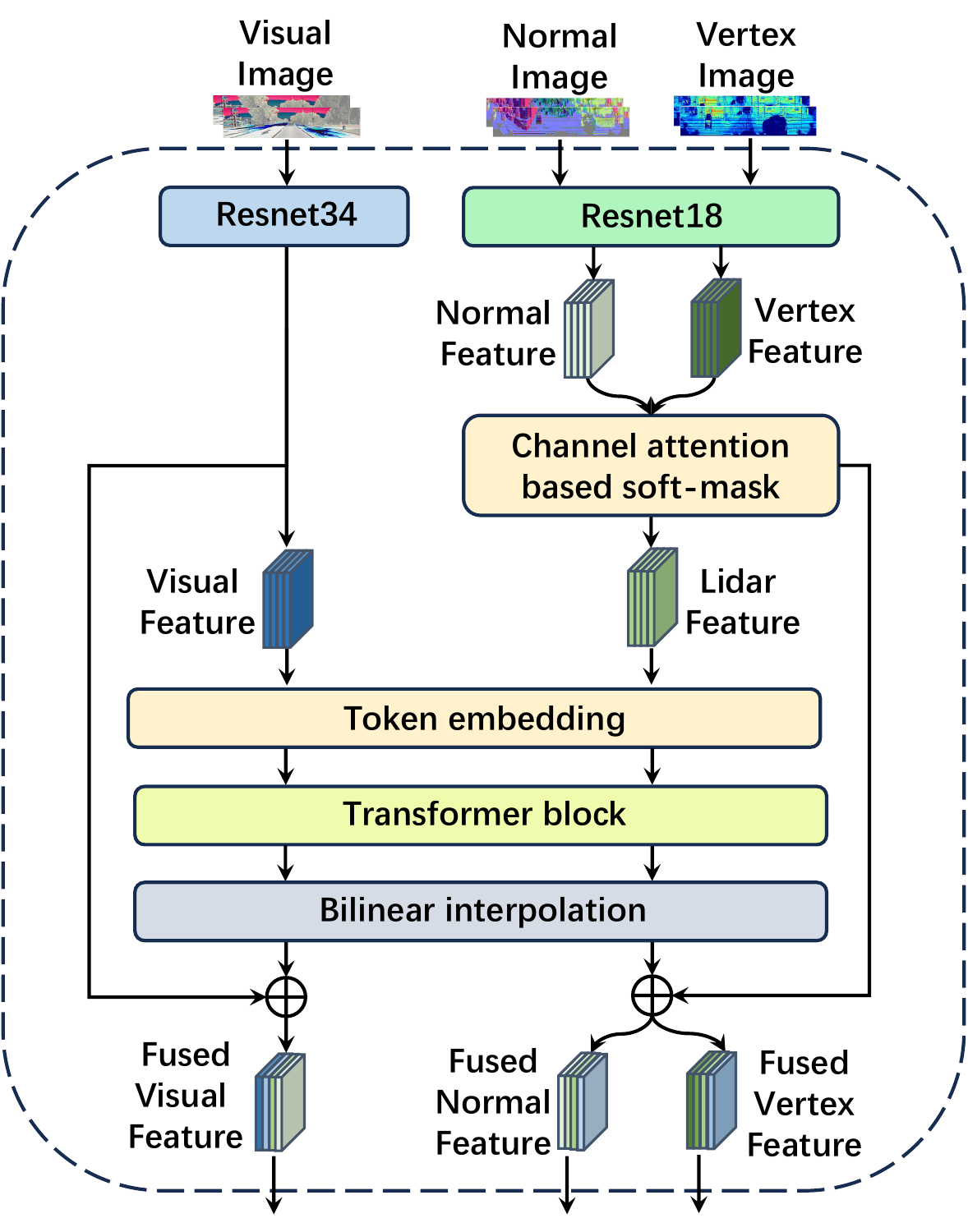}
    \caption{Architecture of the Multi-layer and Multi-scale Image Encoder. The encoder uses ResNet for multi-scale feature extraction and processes them with a Transformer for cross-modal interaction.}
    \label{fig:M-M_Image_encoder}
\end{figure}
We design a multi-layer, multi-scale image feature encoder by integrating ResNet and Transformer architectures to process LiDAR and RGB images. The detailed network architecture is shown in Fig.\ref{fig:M-M_Image_encoder}.

The LiDAR vertex image $\boldsymbol{L}_V$, LiDAR normal image $\boldsymbol{L}_N$, and visual image $\boldsymbol{V}$ are processed through ResNet18 and ResNet34, respectively, to extract compressed multi-scale features $\boldsymbol{L}_V^{l_{i}}$, $\boldsymbol{L}_N^{l_{i}}$ and $\boldsymbol{V}^{l_{i}}$. To reduce complexity, the LiDAR images, originating from the same sensor, share the weights of ResNet18, while the RGB images, being from a different sensor modality, utilize ResNet34 to capture richer texture information. The LiDAR vertex features $\boldsymbol{L}_V^{l_{i}}$ and normal features $\boldsymbol{L}_N^{l_{i}}$ are fused into $\boldsymbol{L}^{l_{i}}$ using a Multilayer
Perceptron(MLP) channel attention-based soft-mask network\cite{chen2019selectfusion}.
The visual features $\boldsymbol{V}^{l_{i}}$ and fused LiDAR features $\boldsymbol{L}^{l_{i}}$ are then embedded into tokens for further processing by the Transformer.
Inspired by Transfuser\cite{chitta2022transfuser}, average pooling is applied to reduce computational complexity by downsampling the original features $\boldsymbol{V}^{l_{i}}$, $\boldsymbol{L}^{l_{i}}$ to $\boldsymbol{V}_s^{l_{i}}$, $\boldsymbol{L}_s^{l_{i}}$, and positional encodings $L^\mathrm{pos}/V^\mathrm{pos}$ are added to retain spatial order. Additionally, modality type encodings $L^\mathrm{type}/V^\mathrm{type}$ are incorporated to differentiate the sensor sources. The embedding process is summarized as:
\begin{equation}
    \begin{aligned}
    &\boldsymbol{\bar{L}}^{l_{i}} = \boldsymbol{L}_s^{l_{i}} + L^\mathrm{pos} \\
    &\boldsymbol{\bar{V}}^{l_{i}} = \boldsymbol{V}_s^{l_{i}} + V^\mathrm{pos} \\ \label{eq:eq5}
    &\boldsymbol{G}^\mathrm{in} =\left[\boldsymbol{\bar{L}}^{l_{i}} + L^\mathrm{type};\boldsymbol{\bar{V}}^{l_{i}} + V^\mathrm{type}\right].
    \end{aligned}
\end{equation}

The Transformer receives an input tensor \(\boldsymbol{G}^{in}\) of dimensions \(N \times D_{f}\), where \(N\) is the token count, and \(D_{f}\) is the feature dimension. The query \(\boldsymbol{Q}\), key \(\boldsymbol{K}\), and value \(\boldsymbol{V}\) are generated through linear transformations of \(\boldsymbol{G}^{in}\) using respective weight matrices $\boldsymbol{M}^{q} \in \mathbb{R}^{D_{f}\times D_{q}}, \boldsymbol{M}^{k} \in \mathbb{R}^{D_{f}\times D_{k}}, \boldsymbol{M}^{v} \in \mathbb{R}^{D_{f}\times D_{v}}$:
\begin{equation}\label{eq:eq6}
\boldsymbol{Q}=\boldsymbol{G}^\mathrm{in}\boldsymbol{M}^q,\quad \boldsymbol{K}=\boldsymbol{G}^\mathrm{in}\boldsymbol{M}^k, \quad \boldsymbol{V}=\boldsymbol{G}^\mathrm{in}\boldsymbol{M}^v .
\end{equation}

Attention scores \(\alpha_{L,V}\) are calculated using scaled dot products of \(\boldsymbol{Q}\) and \(\boldsymbol{K}\), followed by a softmax to derive attention weights. These weights are then used to aggregate \(\boldsymbol{V}\) into the attention output \(\boldsymbol{C}_{L,V}\). The final output features \(\boldsymbol{G}^{\mathrm{out}}\) are computed by applying an MLP to \(\boldsymbol{C}_{L,V}\) and adding the original input features \(\boldsymbol{G}^{in}\):
\begin{equation}
    \begin{aligned}
    &\alpha_{L,V} =\frac{\mathbf{Q}\mathbf{K}^{T}}{\sqrt{D_{k}}} \\
    &\boldsymbol{C}_{L,V} =\mathrm{softmax}(\alpha_{L,V})\boldsymbol{V} \\ \label{eq:eq7}
    &\boldsymbol{G}^{\mathrm{out}} =\mathrm{MLP}(\boldsymbol{C}_{L,V})+\boldsymbol{G}^{\mathrm{in}} .
    \end{aligned}
\end{equation}

The output $\boldsymbol{G}^{\mathrm{out}}$ is then upsampled to its original resolution using bilinear interpolation and added element-wise to the ResNet outputs to enable residual learning, preventing gradient vanishing. This allows ResNet to progressively extract multi-scale features from Vertex image, Normal image, and RGB image, while the Transformer enables effective cross-modal interaction, forming the proposed multi-layer multi-scale image feature encoder.

\subsection{Adaptive Degradation Feature Filter}
\label{filter_section}
To address the complex sensor degradation scenarios in real-world driving conditions, the simple Multilayer Perceptron(MLP)-based reweighting strategy, as proposed in \cite{chen2019selectfusion}, does not yield satisfactory results, while overly complex network structures risk inefficiency and overfitting. We propose a coarse-to-fine temporal and spatial feature filter strategy to balance efficiency, robustness, and accuracy.
Initially, the features at time $t$ undergo coarse temporal filtering using a multi-head attention network to remove redundant temporal features, similar to keyframe extraction in traditional SLAM. Subsequently, the concatenated features are further refined through spatial filtering using a self-attention mechanism.

\begin{figure}[t]
    \centering
    \includegraphics[width=\linewidth]{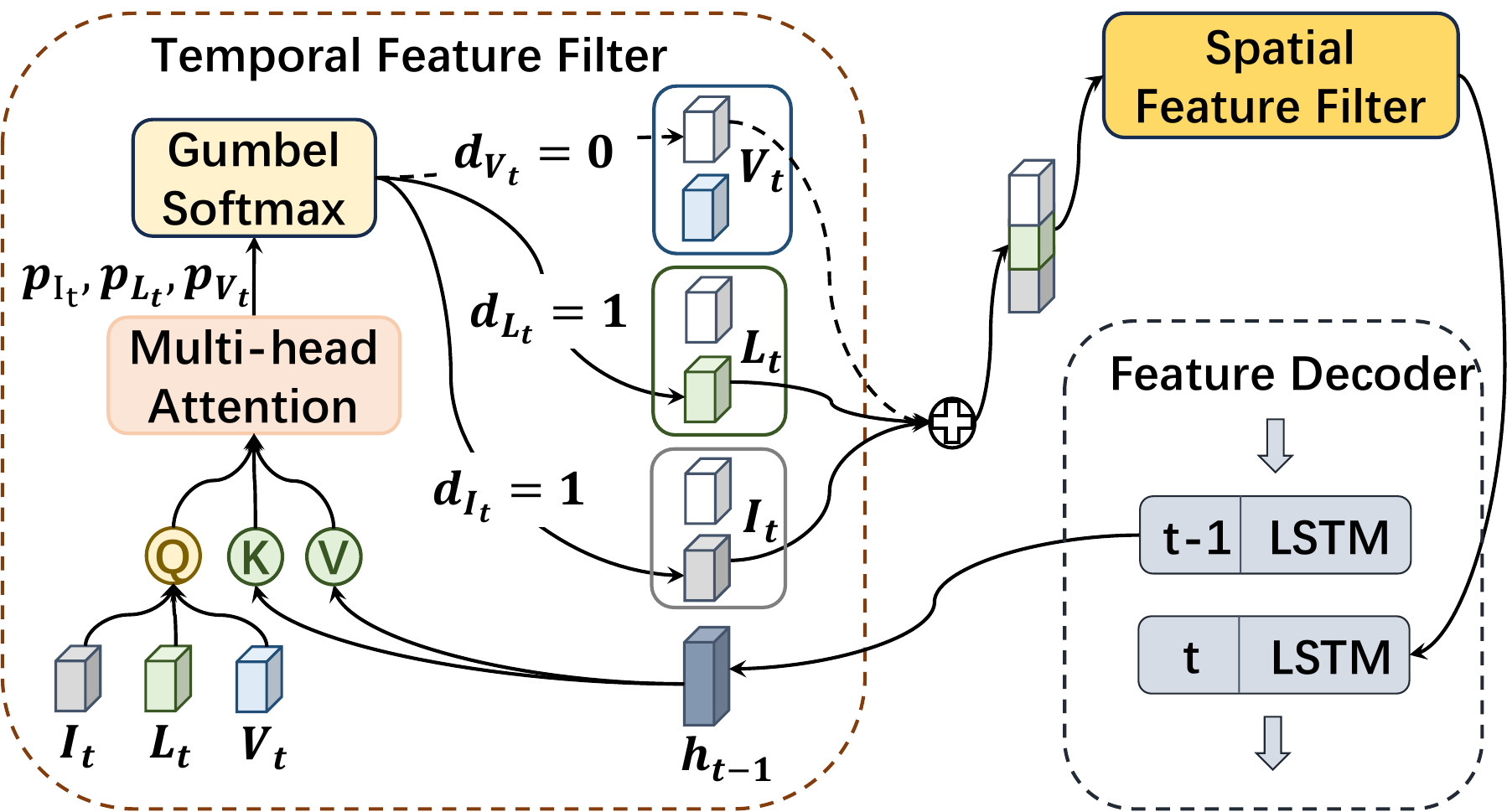}
    \caption{Temporal Feature Filter.  }
    \label{fig:Temp_filter}
\end{figure}
\subsubsection{Temporal Feature Filter} 
As illustrated in Fig.\ref{fig:Temp_filter}, the hidden state \(\boldsymbol{h}_{t-1}\) from the Feature Decoder at the previous time step serves as the key \(\boldsymbol{K}\) and value \(\boldsymbol{V}\), while the LiDAR \(\boldsymbol{L}_t\), visual \(\boldsymbol{V}_t\), and IMU \(\boldsymbol{I}_t\) features at time \(t\) act as the query \(\boldsymbol{Q}\). These inputs are processed by a Multi-Head Attention (MHA) network, where each feature at time \(t\) queries the hidden state \(\boldsymbol{h}_{t-1}\), generating the probabilities \(p_{\boldsymbol{V}_t}, p_{\boldsymbol{L}_t}, p_{\boldsymbol{I}_t}\in\mathbb{R}^{2}\), 
which indicate whether to discard the current frame:
\begin{equation} \label{eq:eq10}
    \begin{aligned}
        &p_{\boldsymbol{V}_t} = \mathrm{MHA}(\boldsymbol{Q} = \boldsymbol{V}_t, \boldsymbol{K} = \boldsymbol{h}_{t-1}, \boldsymbol{V} = \boldsymbol{h}_{t-1}) \\
        &p_{\boldsymbol{L}_t} = \mathrm{MHA}(\boldsymbol{Q} = \boldsymbol{L}_t, \boldsymbol{K} = \boldsymbol{h}_{t-1}, \boldsymbol{V} = \boldsymbol{h}_{t-1}) \\
        &p_{\boldsymbol{I}_t} = \mathrm{MHA}(\boldsymbol{Q} = \boldsymbol{I}_t, \boldsymbol{K} = \boldsymbol{h}_{t-1}, \boldsymbol{V} = \boldsymbol{h}_{t-1}).
    \end{aligned}
\end{equation}

The decision to discard a frame is made using Gumbel-Softmax re-sampling to ensure differentiability during training, following \cite{jang2016categorical}. Decision variables \(d_{\boldsymbol{V}_t}, d_{\boldsymbol{L}_t}, d_{\boldsymbol{I}_t} \in \{0,1\}\) are sampled as \(d_{\boldsymbol{V}_t} \sim \mathrm{GUMBEL}(p_{\boldsymbol{V}_t})\), \(d_{\boldsymbol{L}_t} \sim \mathrm{GUMBEL}(p_{\boldsymbol{L}_t})\), and \(d_{\boldsymbol{I}_t} \sim \mathrm{GUMBEL}(p_{\boldsymbol{I}_t})\). When \(d_{\boldsymbol{V}_t} = 1\), \(d_{\boldsymbol{L}_t} = 1\), and \(d_{\boldsymbol{I}_t} = 1\), the respective feature is retained for further processing; otherwise, it is discarded. The temporally filtered feature vector \(\boldsymbol{F}_t\) at time \(t\) is obtained by concatenating the retained components of \(\boldsymbol{L}_t\), \(\boldsymbol{V}_t\), and \(\boldsymbol{I}_t\), as described by the equation below, where \(\oplus\) denotes concatenation:
\begin{equation} \label{eq:eq12}
    \boldsymbol{F}_t = (d_{\boldsymbol{V}_t} \cdot \boldsymbol{v}_t) \oplus (d_{\boldsymbol{L}_t} \cdot \boldsymbol{L}_t) \oplus (d_{\boldsymbol{I}_t} \cdot \boldsymbol{I}_t).
\end{equation}

\subsubsection{Spatial Feature Filter}
\begin{figure}[t]
    \centering
     \includegraphics[width=\linewidth]{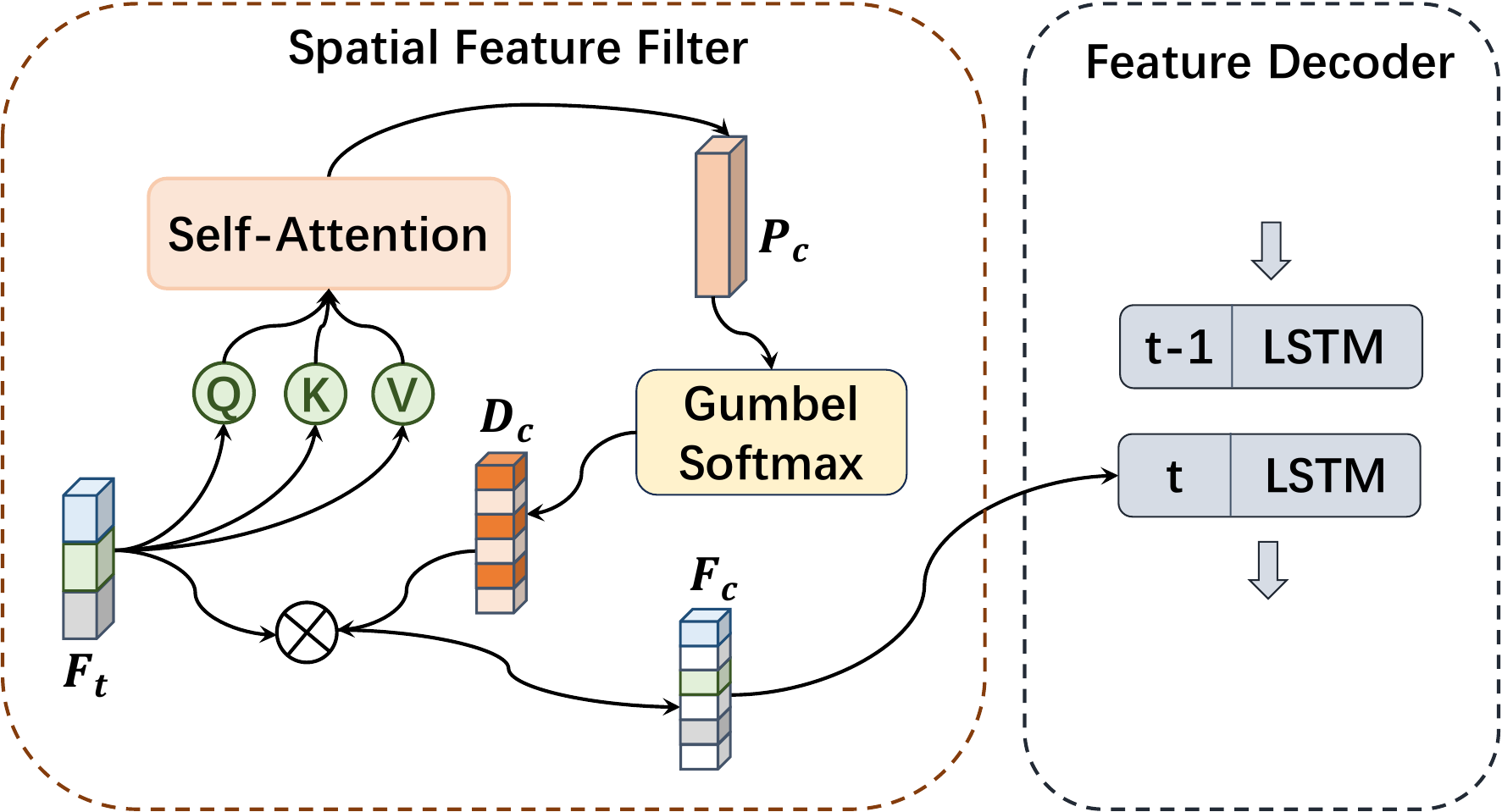}
    \caption{Spatial Feature Filter. }
    \label{fig:Spatial_filter}
\end{figure}

Following coarse temporal filtering, spatial features $\boldsymbol{F}_t$ undergo further refinement using Self-Attention(SA), as shown in Fig.\ref{fig:Spatial_filter}. The query $\boldsymbol{Q}$, key $\boldsymbol{K}$, and value $\boldsymbol{V}$ are all set to $\boldsymbol{F}_t$. The output is a probability $\boldsymbol{P}_c$ for each feature channel, representing whether to retain or discard specific channels. This decision is made using Gumbel-Softmax re-sampling, resulting in the fine-filtered features $\boldsymbol{F}_c$. The detailed process is articulated by the following equations, where \(\otimes\) denotes element-wise multiplication:
\begin{equation}
    \begin{aligned}
    &\boldsymbol{P}_c=\mathrm{SA}(\boldsymbol{Q}=\boldsymbol{F}_t, \boldsymbol{K}=\boldsymbol{F}_t, \boldsymbol{V}=\boldsymbol{F}_t) \\
    &\boldsymbol{D}_c\sim\mathrm{GUMBEL}(\boldsymbol{P}_c) \\ \label{eq:eq15}
    &\boldsymbol{F}_c=\boldsymbol{F}_t\otimes \boldsymbol{D}_c .
    \end{aligned}
\end{equation} 

\subsection{Loss function } 
We design the loss function to balance relative motion between consecutive frames and the global trajectory error. To address differences in units and scales between translation and rotation, we use the homoscedastic weighted sum loss \cite{kendall2018multi}, which introduces learnable task-balancing parameters. The loss function is defined as:
\begin{align} \label{eq:eq42}
    \mathcal{L}(\theta, s_1, s_2 | X) = \frac{1}{S - 1}\sum_{t=1}^{S - 1} \Big( (\mathcal{L}^{l}_{p,t} + \mathcal{L}^{g}_{p,t}) e^{-s_1} + s_1\\
    + \nonumber (\mathcal{L}^{l}_{r,t} + \mathcal{L}^{g}_{r,t}) e^{-s_2} + s_2 \Big) .
\end{align}
where \(s_1\) and \(s_2\) are learnable parameters representing the predicted uncertainty for position and orientation, respectively. \(\theta\) represents the network learning the balance parameters, and \(X\) denotes the network inputs. \(S\) is the sequence length. \(\mathcal{L}^{l}_{p,t}\) and \(\mathcal{L}^{g}_{p,t}\) refer to local and global position losses, while \(\mathcal{L}^{l}_{r,t}\) and \(\mathcal{L}^{g}_{r,t}\) address local and global rotation losses.

To optimize frame selection decisions \(d_{\boldsymbol{V}_t}, d_{\boldsymbol{L}_t}, d_{\boldsymbol{I}_t}\) in the adaptive temporal filtering process, we introduce a feature usage penalty loss \(\mathcal{L}_{usage}\), which penalizes the over-utilization of feature frames. The penalty is controlled by hyperparameters \(\lambda_{\boldsymbol{V}_t}, \lambda_{\boldsymbol{L}_t}, \lambda_{\boldsymbol{L}_t}\) and is calculated as:
\begin{equation} \label{eq:eq43}
    \mathcal{L}_{usage}=\frac{1}{S - 1}\sum_{t=1}^{S - 1}(\lambda_{\boldsymbol{V}_t} d_{\boldsymbol{V}_t} + \lambda_{\boldsymbol{L}_t} d_{\boldsymbol{L}_t} + \lambda_{\boldsymbol{I}_t} d_{\boldsymbol{I}_t}) .
\end{equation}

The total loss function combines the weighted sum loss and the feature usage penalty:
\begin{equation} \label{eq:eq44} 
    \mathcal{L} = \mathcal{L}(\theta, s_1, s_2 | X) + \mathcal{L}_{usage} .
\end{equation}

\section{EXPERIMENTS}

\subsection{Experimental Setups} \label{test_setups}
\subsubsection{Dataset and Evaluation Metrics}
We evaluate the proposed localization algorithm on three diverse datasets: CARLA-Loc\cite{han2023carla}, KITTI Odometry\cite{geiger2013vision}, and Snail-Radar\cite{huai2024snail}. The performance is assessed using the EVO evaluation tool\cite{grupp2017evo}, which computes the Root Mean Square Error (RMSE) of Absolute Pose Error (APE) and Relative Pose Error (RPE).
\begin{itemize}
    \item CARLA-Loc: A simulated dataset with 7 maps and 42 sequences, captured with multiple sensors under diverse degradation conditions. We use 6 test sequences from map 05 and the remaining for training.
    \item KITTI Odometry: A standard benchmark with sequences 00, 01, 02, 04, 06, 08, 09 for training and 05, 07, and 10 for testing, providing raw data from camera, LiDAR, and IMU, along with ground-truth poses.
    \item Snail-Radar: A real-world dataset with challenging scenarios (night driving, dynamic obstacles, adverse weather) used to evaluate generalization to complex scenarios. 
\end{itemize}

\subsubsection{ Implementation Details}
All models are trained on a server with four NVIDIA RTX 4090 GPUs. The odometry model uses the Adam optimizer with an initial learning rate of 1e-3. The batch size is set to 32, with a sequence length of 11. 
A two-phase training approach is employed to ensure the stable convergence: warm-up phase with a fixed frame rejection probability of 50\%, followed by joint training using Gumbel-Softmax sampling with temperature decay.
The model is pre-trained on the CARLA-Loc synthetic dataset for 100 epochs, consisting of 40 epochs of warm-up training and 60 epochs of joint training. Further, the model undergoes 50 epochs of transfer training on the respective training sets of the real-world KITTI Odometry and Snail-Radar datasets, comprising 5 epochs of warm-up training followed by 45 epochs of joint training.
To assess the models' efficiency for real-time inference on resource-constrained hardware, we conducted benchmarks on an NVIDIA RTX 3060TI GPU, achieving a real-time inference speed of 40-50 frames per second (FPS).

\subsection{Pre-training Evaluation}
\begin{table*}[]
    \centering
    \caption{Absolute Pose Error (APE, Unit m) results on Map 05 in the CARLA-Loc Dataset. }
    \label{tab:pretrain_performance}
    \begin{tabular}{cc|ccc|ccc}
    \hline
    \multirow{2}{*}{Method} & \multirow{2}{*}{Type} & \multicolumn{3}{c|}{Static} & \multicolumn{3}{c}{Dynamic} \\
    ~ & ~ & Clear Noon & Foggy Noon & Rainy Night & Clear Noon & Foggy Noon & Rainy Night \\
    \hline
    ORB3-SVIO\cite{campos2021orb} & VIO & 3.24 & 23.52 & 18.03 & \underline{2.29} & 555.48 & 425.74 \\
    VINS-SVIO\cite{qin2019general} & VIO & 4.03 & fail & fail & 3.97 & fail & 6.76 \\
    \hline
    ALOAM\cite{zhang2014loam} & LO & 4.53* & 4.53* & 4.53* & 93.64* & 93.64* & 93.64* \\
    FASTLIO2\cite{xu2022fast} & LIO & 2.36* & 2.36* & \underline{2.36*} & 2.70* & 2.70* & \underline{2.70*} \\
    \hline
    \textbf{Our A2DO-VIO} & VIO & \underline{2.23} & \underline{2.21} & 4.42 & 3.04 & \underline{1.96} & 3.55 \\
    \textbf{Our A2DO-LIO} & LIO & 2.88* & 2.88* & 2.88* & 4.06* & 4.06* & 4.06* \\
    \textbf{Our A2DO-LVIO} & LVIO & \textbf{0.34} & \textbf{0.34} & \textbf{0.65} & \textbf{0.94} & \textbf{0.77} & \textbf{1.91} \\
    \hline
    \end{tabular}
    \begin{tablenotes}
    \footnotesize
    \item  * : The dataset simulates only static and dynamic LiDAR scenarios. 
    \end{tablenotes}
\end{table*}
Pretraining is conducted on the CARLA-Loc dataset to improve the proposed adaptive odometry's performance in degraded scenarios. Tab.\ref{tab:pretrain_performance} shows that our A2DO-LVIO method achieved the lowest Absolute Pose Error (APE) in translation across all conditions, surpassing traditional methods such as ORB3-SLAM3\cite{campos2021orb} stereo VIO, VINS-Fusion\cite{qin2019general} stereo VIO, ALOAM\cite{zhang2014loam}, and FASTLIO2\cite{xu2022fast}, particularly in challenging foggy and rainy scenarios. Furthermore, our visual system utilizes a single left camera exclusively, underscoring its robustness and superior capability in navigating through degraded scenarios. 

The experimental setup of our system encompasses three distinct configurations: A2DO-VIO (Visual-Inertial Odometry), A2DO-LIO (LiDAR-Inertial Odometry), and A2DO-LVIO (LiDAR-Visual-Inertial Odometry). Both A2DO-VIO and A2DO-LIO configurations exhibited consistent stability across various challenging environments characterized by degraded conditions. Notably, the A2DO-LVIO configuration achieved a marked increase in accuracy, underscoring the efficacy of our proposed multi-scale image feature encoder. This encoder integrates visual and LiDAR data adeptly, leveraging their complementary attributes to enhance the system's localization capabilities significantly.

\subsection{Performance Comparison}
\begin{table}[]
    \centering
    \caption{Average relative translational ($t_{rel} (\%)$) and rotational ($r_{rel} (^\circ)$) error results on KITTI Odometry.}
    \label{table:comp_with_others_table}
    \resizebox{\columnwidth}{!}{
    \begin{tabular}{@{}ccccccc@{}}
    \hline
    Method & Type & Metric & 05 & 07 & 10 & Mean \\
    \hline
    \multirow{2}{*}{VINS-Mono\cite{qin2018vins}} & \multirow{2}{*}{VIO($T$)} & \( t_{rel} (\%) \) & 11.6 & 10.0 & 16.5 & 12.7 \\
    ~ & ~ & \( r_{rel} (^\circ) \) & 1.26 & 1.72 & 2.34 & 1.77 \\
    \multirow{2}{*}{LIO-SAM\cite{shan2020lio}} & \multirow{2}{*}{LIO($T$)} & \( t_{rel} (\%) \) & \underline{1.69} & \underline{2.87} & 4.97 & 3.18 \\
    ~ & ~ & \( r_{rel} (^\circ) \) & 1.28 & 1.62 & 2.17 & 1.69 \\
    \hline
    \multirow{2}{*}{Selectfusion\cite{chen2019selectfusion}} & \multirow{2}{*}{LVO(L)} & \( t_{rel} (\%) \) & 4.25 & 4.46 & 5.81 & 4.84 \\
    ~ & ~ & \( r_{rel} (^\circ) \) & 1.67 & 2.17 & 1.55 & 1.80 \\
    \multirow{2}{*}{ATVIO\cite{liu2021atvio}} & \multirow{2}{*}{VIO($L$)} & \( t_{rel} (\%) \) & 4.93 & 3.78 & 5.71 & 4.81 \\
    ~ & ~ & \( r_{rel} (^\circ) \) & 2.40 & 2.59 & 2.96 & 2.65 \\
    \hline
    \multirow{2}{*}{\textbf{Our A2DO-VIO}} & \multirow{2}{*}{VIO($L$)} & \( t_{rel} (\%) \) & 2.95 & 3.98 & 4.36 & 3.76 \\
     ~ & ~ & \( r_{rel} (^\circ) \) & 1.40 & 2.90 & 1.52 & 1.94 \\
    \multirow{2}{*}{\textbf{Our A2DO-LIO}} & \multirow{2}{*}{LIO($L$)} & \( t_{rel} (\%) \) & 3.84 & 3.21 & 4.80 & 3.95 \\
     ~ & ~ & \( r_{rel} (^\circ) \) & 1.85 & 2.51 & 1.69 & 2.02 \\
     \multirow{2}{*}{\makecell{\textbf{Our A2DO-LVIO} \\ \textbf{(w/o pre-training)}}} & 
     \multirow{2}{*}{LVIO($L$)} & \( t_{rel} (\%) \) & 2.93 & 3.30 & \underline{3.29} & \underline{3.17} \\
     ~ & ~ & \( r_{rel} (^\circ) \) & \underline{0.76} & \underline{1.19} & \underline{0.90} & \underline{0.95} \\
    \multirow{2}{*}{\textbf{Our A2DO-LVIO}} & \multirow{2}{*}{LVIO($L$)} & \( t_{rel} (\%) \) & \textbf{1.24} & \textbf{1.07} & \textbf{1.77} & \textbf{1.36} \\
    ~ & ~ & \( r_{rel} (^\circ) \) & \textbf{0.44} & \textbf{0.67} & \textbf{0.50} & \textbf{0.54} \\
    \hline
    \end{tabular}
    }
    \begin{tablenotes}
    \footnotesize
    \item $T$: Traditional methods. $L$: Learning-based methods.
    \item \textbf{w/o pre-training}: Only 100 epochs training on KITTI Odometry.
    \end{tablenotes}
\end{table}
The proposed adaptive degradation handling odometry is compared using the KITTI Odometry dataset, as detailed in Tab.\ref{table:comp_with_others_table}. Representative methods from traditional and deep learning-based approaches are evaluated, including VINS-Mono\cite{qin2018vins}, LIO-SAM\cite{shan2020lio}, Selectfusion\cite{chen2019selectfusion} and ATVIO \cite{liu2021atvio}. Evaluation metrics, based on average translation and rotation errors, are computed per 100 meters, with all deep learning models trained and tested on specific KITTI sequences.
From the accuracy comparison in the table, the proposed adaptive anti-degradation odometry(A2DO-LVIO), although not specifically designed to enhance localization accuracy but to improve overall robustness, still achieves the best results among all methods. This indicates that degradation scenarios are prevalent in everyday driving conditions, contributing to cumulative errors. Proper handling of these scenarios can enhance both system robustness and accuracy.

\subsection{Ablation Study}
\begin{table}[]
    \centering
    \caption{Ablation Study on Degradation Handling Components.}
    \label{table:ablation_table}
    \begin{tabular}{c|c|cc|c}
    \hline
    \multirow{2}{*}{Method} & \multirow{2}{*}{Type} & \multicolumn{2}{c|}{Static} & Dynamic \\
    ~ & ~ & Clear Noon & Rainy Night & Rainy Night \\
    \hline
    \multirow{2}*{Base} & \( t_{rel} (\%) \) & 2.17 & 3.90 & 5.43 \\
    ~ & \( r_{rel} (^\circ) \) & 2.08 & 1.93 & 2.83  \\
    \hline
    \multirow{2}*{Base+TF} & \( t_{rel} (\%) \) & 1.69 & 2.95 & 3.00  \\
    ~ & \( r_{rel} (^\circ) \) & \underline{0.75} & 0.91 & \underline{0.90}  \\
    \hline
    \multirow{2}*{Base+SF} & \( t_{rel} (\%) \) & \underline{1.67} & \underline{2.40} & \underline{2.56}  \\
    ~ & \( r_{rel} (^\circ) \) & 0.79 & \underline{0.90} & 0.92  \\
    \hline
    \multirow{2}*{A2DO(full)} & \( t_{rel} (\%) \) & \textbf{0.47} & \textbf{0.80} & \textbf{1.24}  \\
    ~ & \( r_{rel} (^\circ) \) & \textbf{0.29} & \textbf{0.35} & \textbf{0.50}  \\
    \hline
    \end{tabular}
\end{table}
To evaluate the effectiveness of the proposed degradation handling mechanisms, ablation studies are conducted on the CARLA-Loc dataset using three scenarios: Static Clear Noon, Static Rainy Night, and Dynamic Rainy Night. As shown in Tab.\ref{table:ablation_table}, 
The Base model exhibits high errors, particularly in the Dynamic Rainy Night scenario, with \( t_{rel} \) at 5.43\% and \( r_{rel} \) at 2.83°, indicating its inability to handle complex scenarios.
Adding the Temporal Feature Filter (TF) in Base+TF significantly reduces errors, especially in Dynamic Rainy Night, where \( t_{rel} \) drops to 3.00\% and \( r_{rel} \) to 0.90°. Similarly, the Spatial Feature Filter (SF) in Base+SF brings further improvements, lowering \( t_{rel} \) to 2.56\% and \( r_{rel} \) to 0.92°.
The full model, A2DO (Base+TF+SF), delivers the best performance, with \( t_{rel} \) at 1.24\% and \( r_{rel} \) at 0.50°, demonstrating the combined effectiveness of both TF and SF in handling degraded scenarios.
Additionally,  Tab.\ref{table:comp_with_others_table} shows that A2DO-LVIO with pre-training on CARLA-Loc outperforms the non-pre-trained version, demonstrating the benefits of pre-training in enhancing localization performance.

\begin{figure}[]
    \centering
        \includegraphics[width=0.7\linewidth]{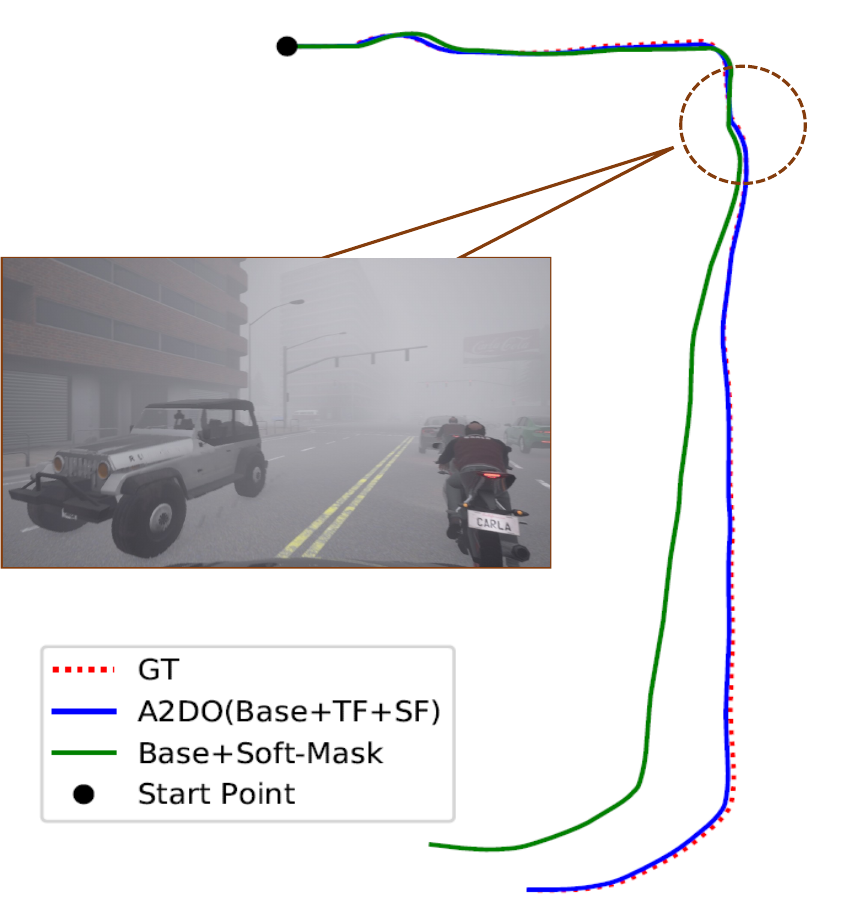}
    \caption{Comparison of A2DO (Base+TF+SF) and Soft-Mask strategies on the map 05 Dynamic Foggy sequences, demonstrating superior robustness of A2DO in challenging conditions.}
    \label{fig:traj_comp_graph}
\end{figure}

Furthermore, Fig.\ref{fig:traj_comp_graph} compares our A2DO (Base+TF+SF) with the Soft-Mask approach from SelectFusion, using the map 05 Dynamic Foggy sequence.The results indicate that our method handles challenging scenarios, such as dense fog and dynamic vehicle occlusions, more robustly, providing stable localization, while the Soft-Mask approach exhibits less stability. This further demonstrates the superiority of our degradation handling strategy.

\subsection{Generalization Ability Verification}
\begin{figure}[]
    \centering
    \includegraphics[width=\linewidth]{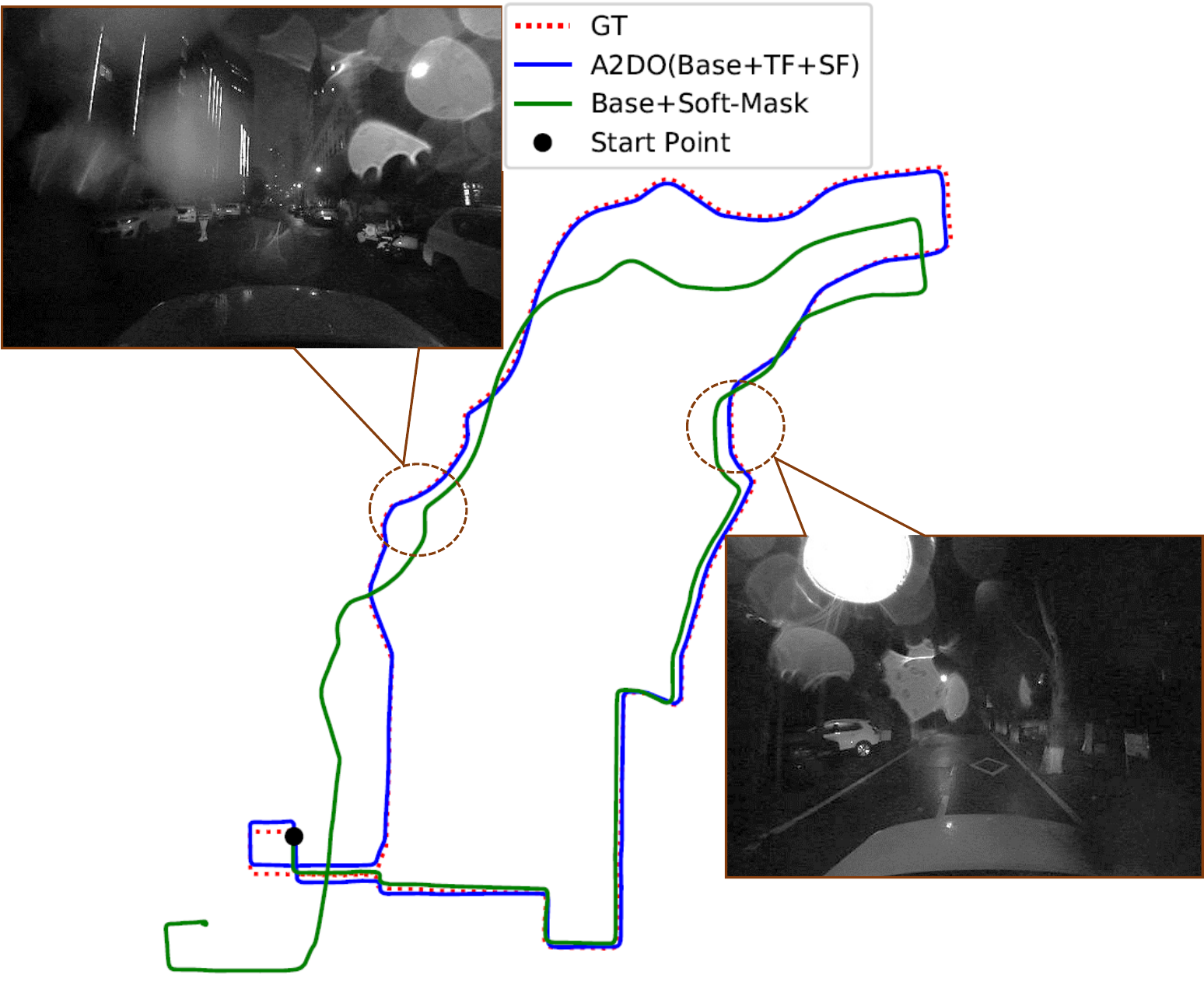}
    \caption{ Trajectories under rainy night conditions on Snail-Radar dataset 20231208 sequence 3, comparing the proposed A2DO with a Soft-Mask-based method.}
    \label{fig:generalization}
\end{figure}

To validate the generalization of the algorithm in real-world driving conditions with degradation scenarios, tests are conducted on the Snail-Radar dataset, and the test setup is the same with \ref{test_setups}. 
The test results show an relative translational error (\( t_{rel} (\%) \)) of 1.82, an relative rotational error (\( r_{rel} (^\circ) \)) of 0.48. These results are comparable to those obtained from the KITTI Odometry dataset and the CARLA-Loc simulation dataset, demonstrating that the proposed algorithm applies to real-world driving conditions.
The overall localization trajectory is shown in Fig.  \ref{fig:generalization}. 
Despite camera occlusions, glare, dynamic objects, and LiDAR noise from raindrops, the proposed A2DO method maintains robust localization, while the Soft-Mask-based method exhibits severe drift, validating the effectiveness of our approach in degraded driving conditions.

\section{CONCLUSIONS}
This paper proposes a robust adaptive anti-degradation multi-sensor fusion localization algorithm to address the issue of sensor degradation. The algorithm incorporates a multi-layer, multi-scale image feature encoder and a coarse-to-fine temporal-spatial hierarchical filtering strategy to fuse multi-modal sensor data and handle degraded conditions effectively. Extensive experimental results demonstrate that the proposed method handles various degraded scenarios with high precision. By leveraging comprehensive pre-training on simulated datasets, the algorithm reduces the reliance on real-world degraded data for transfer learning, achieving robust and accurate localization in real-world driving conditions. 
Future work will focus on developing efficient transfer learning methods for zero-shot learning and exploring cost-effective sensor alternatives to LiDAR.




\bibliographystyle{IEEEtran}
\bibliography{IEEEabrv,a2do}
\end{document}